\documentclass[letterpaper, 10 pt, conference]{ieeeconf}  %

\IEEEoverridecommandlockouts

\usepackage{graphics} %
\usepackage{epsfig} %
\usepackage{times} %
\usepackage{amsmath} %
\usepackage{amssymb}  %
\usepackage{bbm}
\usepackage{booktabs}
\usepackage{mathtools}
\usepackage{multirow}
\usepackage{units}
\usepackage{bm}
\usepackage{xcolor}
\let\labelindent\relax
\usepackage{enumitem}
\usepackage{balance}
\usepackage[ruled,vlined]{algorithm2e}
\usepackage{algorithmicx}
\usepackage{cite}
\usepackage{tabularx}
\usepackage[caption=false, font=footnotesize]{subfig}
\usepackage{graphicx}
\usepackage{adjustbox}
\usepackage{letltxmacro}
\LetLtxMacro{\originaleqref}{\eqref}
\renewcommand{\eqref}{Eq.~\originaleqref}
\usepackage{algpseudocode}
\algnewcommand\Algand{\textbf{and} }

\newcommand{\ie}{\emph{i.e.},}

\SetCommentSty{mycommfont} %
\usepackage{microtype}

\definecolor{LightCyan}{rgb}{0.88,1,0.88}
\definecolor{emb_color}{RGB}{252,224,225}
\definecolor{multi_head_attention_color}{RGB}{252,226,187}
\definecolor{add_norm_color}{RGB}{242,243,193}
\definecolor{ff_color}{RGB}{194,232,247}
\definecolor{softmax_color}{RGB}{203,231,207}
\definecolor{linear_color}{RGB}{220,223,240}
\definecolor{gray_bbox_color}{RGB}{243,243,244}

\usepackage{hyperref}

\usepackage{subcaption}
\usepackage{amsmath}
\usepackage{amssymb}
\usepackage{pifont}
\usepackage{arydshln}
\usepackage{placeins} %
\usepackage{threeparttable}
\usepackage{dirtree}
\usepackage{balance}
\usepackage{multirow}
\usepackage{booktabs}

\usepackage{fancyhdr}
\fancypagestyle{withfooter}{
  
  \fancyfoot[C]{\footnotesize Accepted to the IEEE ICRA Workshop on Field Robotics 2024}
}

\definecolor{mycolor}{RGB}{180, 211, 178}
\definecolor{mycolor_old}{rgb}{0.96,0.87,0.83}

\definecolor{bush}{RGB}{230, 25, 75}
\definecolor{dirt}{RGB}{60, 180, 75}
\definecolor{fence}{RGB}{0, 128, 128}
\definecolor{grass}{RGB}{128, 128, 128}
\definecolor{gravel}{RGB}{145, 30, 180}
\definecolor{log}{RGB}{128, 128, 0}
\definecolor{mud}{RGB}{255, 225, 25}
\definecolor{object}{RGB}{250, 190, 190}
\definecolor{other-terrain}{RGB}{70, 240, 240}
\definecolor{rock}{RGB}{170, 255, 195}
\definecolor{sky}{RGB}{0, 0, 128}
\definecolor{structure}{RGB}{170, 110, 40}
\definecolor{tree-foliage}{RGB}{210, 245, 60}
\definecolor{tree-trunk}{RGB}{240, 50, 230}
\definecolor{water}{RGB}{0, 130, 200}

\begin{document}

\title{
\LARGE \bf
Towards Long-term Robotics in the Wild

}
\author{Stephen Hausler$^{*1}$, Ethan Griffiths$^{*1,2}$, Milad Ramezani$^{1}$, Peyman Moghadam$^{1,2}$
\thanks{$^*$ Equally contributed authors}
\thanks{$^1$ Authors are with the CSIRO Robotics, DATA61, CSIRO, Brisbane, QLD 4069, Australia. 
E-mails: {\tt\footnotesize \emph{firstname.lastname}@csiro.au}}
\thanks{
$^{2}$  Authors are with the School of Electrical Engineering and Robotics, Queensland University of Technology (QUT), Brisbane, Australia.
E-mails: {\tt\footnotesize \emph{firstname.lastname}@qut.edu.au}
}
} 

\bstctlcite{IEEEexample:BSTcontrol}

\maketitle

\thispagestyle{withfooter}
\pagestyle{withfooter}

\begin{abstract}

In this paper, we emphasise the critical importance of large-scale datasets for advancing field robotics capabilities, particularly in natural environments. While numerous datasets exist for urban and suburban settings, those tailored to natural environments are scarce. Our recent benchmarks WildPlaces and WildScenes address this gap by providing synchronised image, lidar, semantic and accurate 6-DoF pose information in forest-type environments. We highlight the multi-modal nature of this dataset and discuss and demonstrate its utility in various downstream tasks, such as place recognition and 2D and 3D semantic segmentation tasks.

\end{abstract}

\section{Introduction}

In the era of large models, the need for large-scale benchmarks has become increasingly crucial, especially for field mobile robots operating in natural, unstructured environments. The complexity of natural environments, characterised by their irregular terrain, dense vegetation, unstructured elements, and the dynamic shifts of such environments, presents a significant challenge to current autonomous agents\cite{frey2023fast,ramezani2023deep, baril2021kilometer,lowe2021canopy}. These agents must autonomously localise, map, and understand these complex environments to function effectively. 

While substantial progress has been made in establishing benchmarks for urban and on-road settings, exemplified by datasets like KITTI360 \cite{liao2022kitti360}, Argoverse \cite{chang2019argoverse}, and Boreas \cite{burnett2023boreas}, benchmarks for natural, unstructured environments still significantly lag behind.

Besides, the recent trends of scaling up model sizes have significantly improved the model generalisability (urban-to-urban) in various important downstream tasks.  However, the performance of these models degrades significantly in the presence of severe environmental domain shifts such as from urban to natural environments \cite{geoadapt2024}. 

A key challenge towards establishing benchmarks in natural environments is to obtain accurate localisation to facilitate research on long-term robotics such as inter-sequence tasks (\ie{} re-localisation). Natural environments, in particular forest trails, cause perturbations to the global navigation satellite system (GNSS) signals due to interference of the dense canopy. Prior work \cite{baril2021kilometer} reported up to $8m$ GNSS errors in forest trails. 

To address these challenges, we introduce WildPlaces\cite{knights2023wildplaces} and WildScenes\cite{vidanapathirana2023wildscenes}, large-scale benchmarks in natural forest trails for intra-sequence and inter-sequence lidar place recognition tasks, and long-term 2D, 3D semantic segmentation tasks, respectively. We also provide accurate 6-DoF ground truth poses using our lidar-inertial Wildcat SLAM~\cite{ramezani2022wildcat} to further facilitate research in long-term robotics. 

The dataset was collected by walking through dense forest trails over the period of 14 months with a portable, handheld sensor payload. The sensor payload comprises rich input modalities of 3D lidar, RGB images, IMU, and GPS with manually annotated 2D semantic segmentation images, generated 3D semantic segmentation point clouds, coupled with accurate 6-DoF ground truth pose. 

In this paper, we first provide an overview of the benchmarks. Then, we provide a summary of established baselines for inter-sequence and intra-sequence lidar place recognition, multi-modal place recognition, 2D semantic segmentation, 3D semantic segmentation, and 2D and 3D domain shifts across environmental and temporal domains.

\section{Background and Related Work}

To enhance robotics capabilities in natural environments, a variety of datasets have been curated and tailored to applications in field robotics\cite{liu2024botanicgarden, baril2021kilometer, jiang2021rellis, min2022orfd, wigness2019rugd, triest2022tartandrive}. These datasets offer a rich diversity of information, including RGB images, semantic annotations, and lidar point clouds. Notable datasets in this domain include RELLIS-3D\cite{jiang2021rellis}, ORFD\cite{min2022orfd}, RUGD\cite{wigness2019rugd} and TartenDrive\cite{triest2022tartandrive}, each providing unique insights into off-road scenarios and forest environments.
RELLIS-3D\cite{jiang2021rellis} stands out for its provision of both RGB and lidar annotations, facilitating semantic segmentation tasks in natural landscapes. Meanwhile, ORFD\cite{min2022orfd} offers a comprehensive dataset for free-space and traversability detection, featuring semantic labels for delineating various terrain types in off-road scenarios under various weather conditions. However, its semantic labels consist of only three classes: free space, traversable and non-traversable. 
RUGD\cite{wigness2019rugd}, on the other hand, contributes a large-scale dataset captured in off-road terrains, focusing primarily on RGB imagery. 
Datasets such as TartenDrive\cite{triest2022tartandrive} focus on learning off-road vehicle dynamics. 
Additionally, \cite{valada16iser} offers Freiburg Forest, a multi-spectral segmentation benchmark, and demonstrates that other spectra in combination with RGB channels improve segmentation. 

The datasets mentioned above serve as invaluable resources for training and evaluating robotics systems designed to navigate and operate in challenging natural environments. By leveraging the diverse modalities and rich annotations provided by these datasets, one can develop and validate robust algorithms for tasks such as semantic segmentation, object detection, and traversability analysis. 
However, these datasets focus only on off-road environments suitable for ground vehicles. Exploring a multi-modal dataset with segmentation benchmarking in 2D and 3D in dense forest environments contributes to the navigation advancement of robot platforms, including legged and flying robots under the canopy.
\begin{figure}[t]
    \centering
    \includegraphics[width=0.99\columnwidth, trim=0cm 2cm 1.5cm 0.0cm,clip]{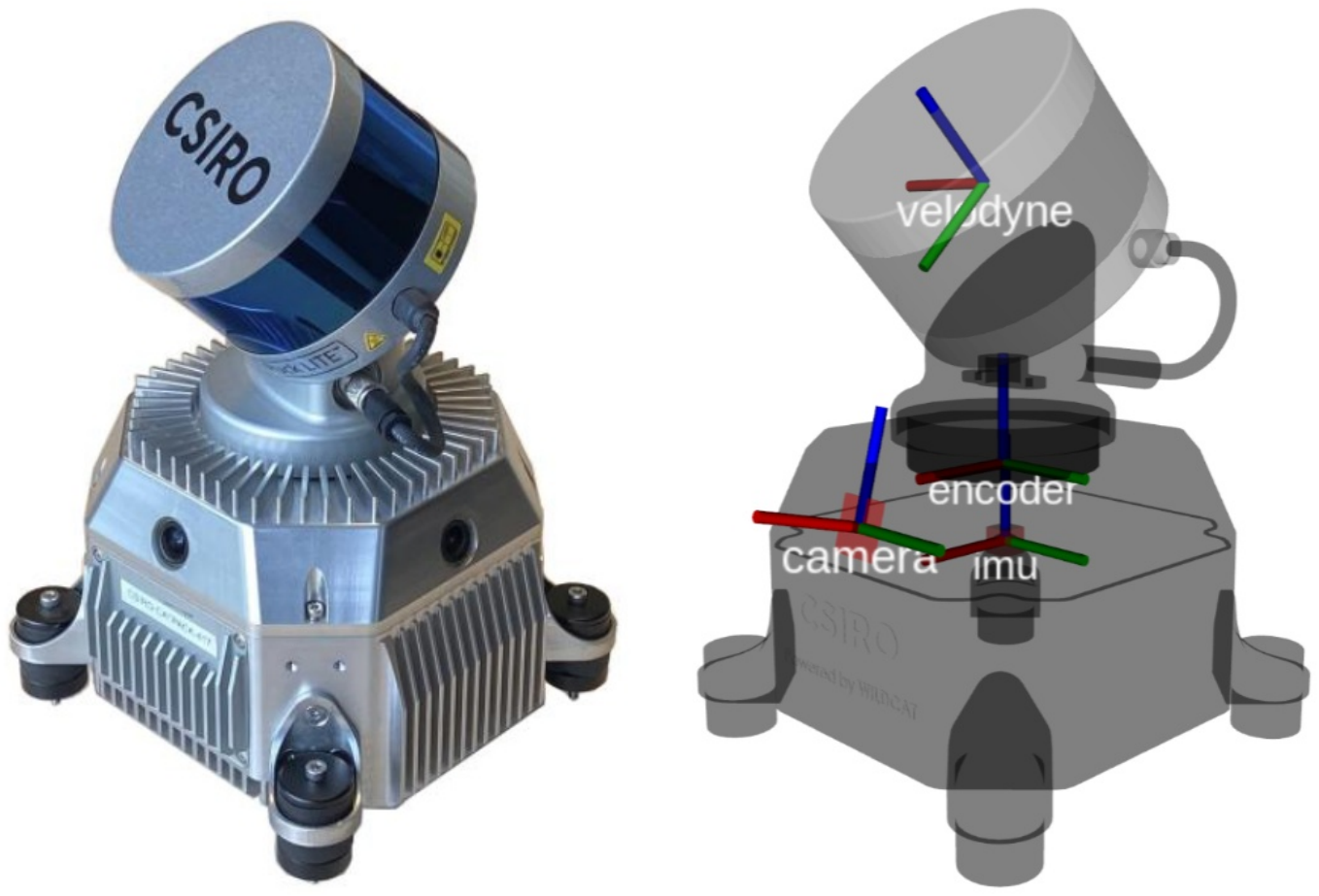}
    \caption{Our sensor platform is a modular device which contains four cameras, a spinning lidar sensor, encoder, IMU and GPS, and can be hand-held or mounted on a mobile robot.
    }
    \label{fig:datacollection}
\end{figure}
\begin{figure*}[t]
    \centering
    \includegraphics[width=0.99\linewidth, trim=0cm 0.5cm 4cm 0.0cm,clip]{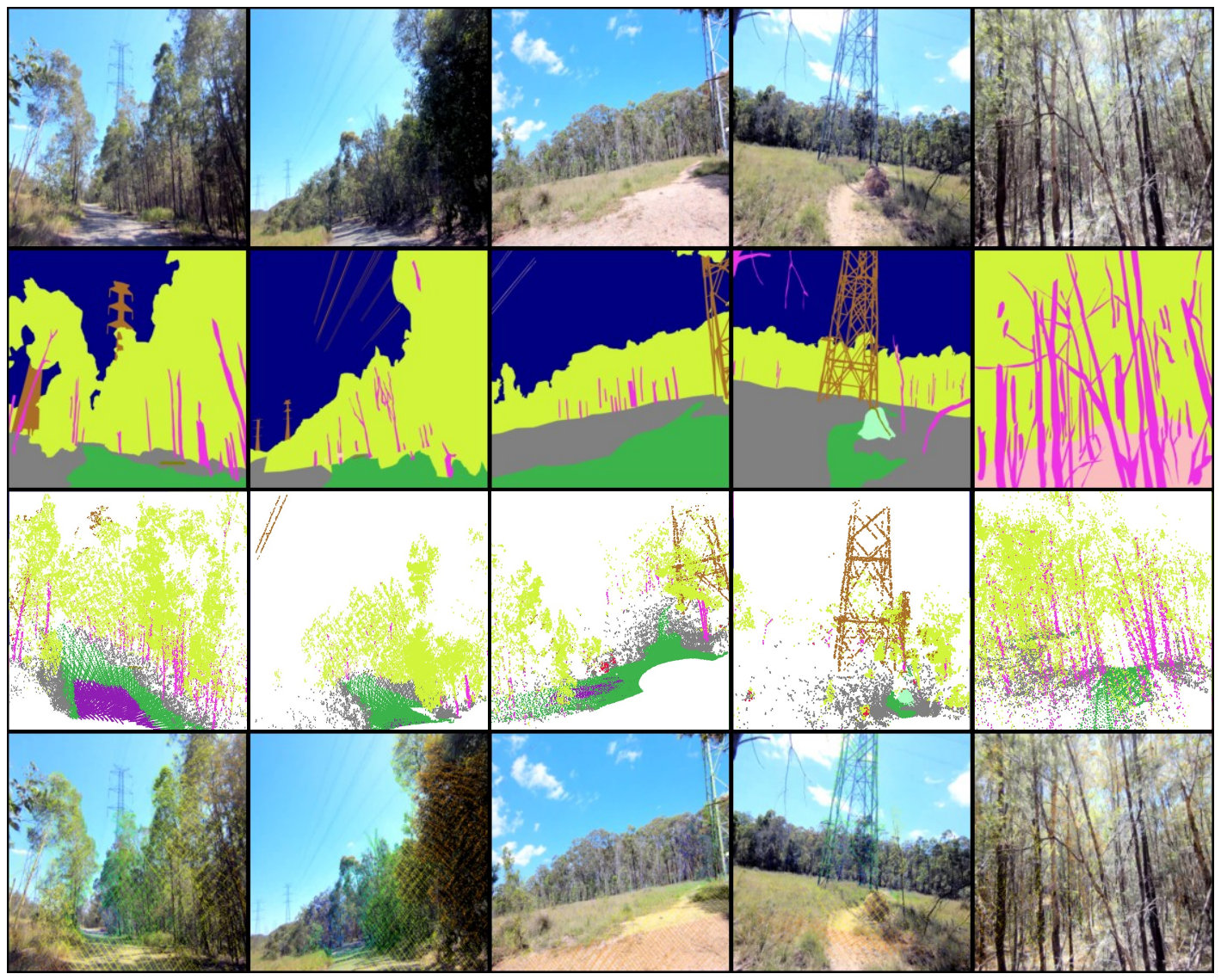}
    \caption{Example places from our dataset showing the different modalities. The first row shows the RGB modality, the second row the 2D annotated image, row three shows the 3D point cloud which has also been annotated with semantics. The bottom row displays the projection from 3D onto the 2D images, colour coded by depth.
    }
    \label{fig:qual}
\end{figure*}

\section{Dataset Description}

The dataset used for WildPlaces~\cite{knights2023wildplaces} and WildScenes~\cite{vidanapathirana2023wildscenes} is a multi-modal dataset collected from natural forest environments in Brisbane, Australia. It includes eight traverses through two forests, Venman and Karawatha, conducted between June 2021 and August 2022 (over 14 months) in multiple sessions with slightly varying trajectories (including reverse revisits). This dataset contains multiple revisits both within and between sequences. The data was collected by hiking with a portable, handheld sensor payload, as shown in Figure~\ref{fig:datacollection}, covering approximately 33km.

WildPlaces~\cite{knights2023wildplaces} establishes benchmarks for lidar place recognition both intra-sequence (\ie{} loop closure detection) and inter-sequence (\ie{} re-localisation) tasks. 
WildScenes~\cite{vidanapathirana2023wildscenes}, successor of WildPlaces~\cite{knights2023wildplaces}, includes additional layers of scene information suitable for 2D and 3D semantic segmentation tasks. WildScenes also includes baselines for domain adaptation with both environmental and temporal domain shifts. 
These benchmarks have separate pre-processing steps and data curation steps in order to maximise the utility of the underlying raw data for these specific tasks. 

In the following subsections, we briefly recap our sensor platform, ground truth setup, pre-processing procedures, and semantic annotations. For further details, please see\cite{knights2023wildplaces, vidanapathirana2023wildscenes}. In summary, the full data collection consists of images, point clouds, 2D semantic annotated images, 3D semantic annotated point clouds, and 6-DoF ground-truth poses.

\subsection{Sensor Platform}

Our portable sensor setup includes a Velodyne Puck lidar sensor with 16 beams mounted on a rotating brushless DC motor. This setup provides a 120° vertical Field of View (FoV), ideal for mapping features like trees. Additionally, it features a Microstrain 3DM-CV5-25 9-DoF IMU, a Ublox GPS antenna, and a Nvidia Jetson AGX Xavier. Pulse Per Second (PPS) ensures precise time synchronisation among the sensors. The setup also includes four cameras for visual perception, although only the front camera is used for annotation.

\subsection{Ground Truth}

To provide an accurate localisation and mapping ground truth, we employed Wildcat\cite{ramezani2022wildcat}, a lidar SLAM system based on a Continuous-Time (CT) trajectory optimisation\cite{bosse2009continuous,furgale2012continuous,droeschel2018icra, park2021elasticity}. Benefiting from CT trajectory estimation and incorporating lidar, IMU and GPS measurements, Wildcat corrects lidar scan distortion caused by motion. Additionally, by detecting loop-closures in revisit areas, Wildcat eventually generates a near-ground-truth trajectory and globally consistent undistorted map of the environment. To enable multi-session place recognition, we further aligned maps of Karawatha and Venman sequences using ICP while removing outliers, resulting in an error of less than $1m$ for above $95\%$ of corresponding points. More information can be found in~\cite{knights2023wildplaces}.

\subsection{Pre-processing}

Our dataset was collected using our sensor platform running Wildcat SLAM and Robot Operating System (ROS). We generated global point clouds per traverse. We then split the global point cloud into multiple submaps, each with a maximum diameter of $60$ meters in WildPlaces and $90$ meters in WildScenes. 
In WildPlaces, each submap comprises almost all lidar points collected by the sensor, resulting in approximately 300K points per submap. However, in WildScenes, to support multi-modality research, we only kept points which are also annotated in 3D, resulting in approximately 70K points per submap.

To extract images, we sub-sampled the original video stream recorded at 15Hz to collect an image for every five meters travelled or after every five degrees of cumulative rotation of the heading angle. We use the forward-facing camera of the four cameras available on the platform. To support multi-modality research, we guarantee that we also generate a corresponding submap at the identical timestamp for every image. By using highly calibrated camera extrinsic and intrinsic parameters, we are able to accurately project lidar points onto the corresponding image.

In WildScenes, we also provided semantic annotations in 2D and 3D, with classes including vegetation categories such as \emph{tree-foliage} for leaves and \emph{tree trunk} for trunks and large branches. Terrain features were also classified, such as \emph{dirt} and \emph{mud} - these are especially useful for downstream tasks such as terrain traversability. The full class list can be found in WildScenes~\cite{vidanapathirana2023wildscenes}.

\section{Dataset Use Cases}

In this section we summarise the results of existing tasks on the WildPlaces and WildScenes benchmarks prior to introducing a multi-modality experiment.

\textbf{2D and 3D semantic segmentation:} We observed that the WildScenes~\cite{vidanapathirana2023wildscenes} is challenging for existing 2D and 3D segmentation baseline algorithms. In the 2D benchmark, DeepLabv3 achieves a mIoU of $42.9\%$ on the WildScenes test set. Per class, the IoU scores differ greatly with the highest IoU of $85.8\%$ achieved by the tree-foliage class. In the 3D benchmark, MinkUNet achieves a mIoU of $34.7\%$, with the highest IoU of $85.7\%$ also achieved by the tree-foliage class.

When a semantic segmentation network is trained on a particular domain of WildScenes and tested on a different one (domain adaptation task), more insights are obtained. Without long-term temporal shifts, DeepLabv3 achieves of a mIoU of $48.5\%$ and MinkUNet achieves a mIoU of $30.3\%$. With temporal changes (Summer to Winter), DeepLabv3 reaches a mIoU of $44.0\%$ and MinkUNet reaches a mIoU of $27.2\%$. This reduction in performance is an expected drop due to the change in appearance that occurs over time.

\textbf{Lidar place recognition:} lidar place recognition was evaluated using the point cloud submaps in WildPlaces~\cite{knights2023wildplaces}. These submaps have a 360-degree field of view and contain 300K points on average. Considering inter-sequence place recognition, LoGG3D-Net\cite{vid2022logg3d} achieves a recall@1 of $79.8\%$ on the Venman sequences and a recall@1 of $74.7\%$ on the Karawatha sequences, while MinkLoc3Dv2\cite{komorowski2022improving} achieves a recall@1 of $75.8\%$ on Venman and a recall@1 of $67.8\%$ on Karawatha. Similar to the semantic segmentation task, each model is first trained using a training split and then evaluated on a disjoint test set. 

For the application of long-term robotics, WildPlaces also ran experiments specifically to test the impact of temporal changes on place recognition performance. In the Venman environment, when the query and database sets are from two sessions captured on the same day, the recall@1 for LoGG3D-Net and MinkLoc3Dv2 is approximately $90\%$ for both models. When the query set is instead collected 14 months later, the recall@1 for LoGG3D-Net and MinkLoc3Dv2 dropped to approximately $74\%$ and $70\%$, respectively. Therefore, as was the case for the segmentation task, long-term robotics tasks still pose potential difficulties due to the natural appearance changes over time.

\begin{figure}[t]
    \centering
    \includegraphics[width=1.0\linewidth, trim=10.1cm 5cm 6.5cm 2.5cm,clip]{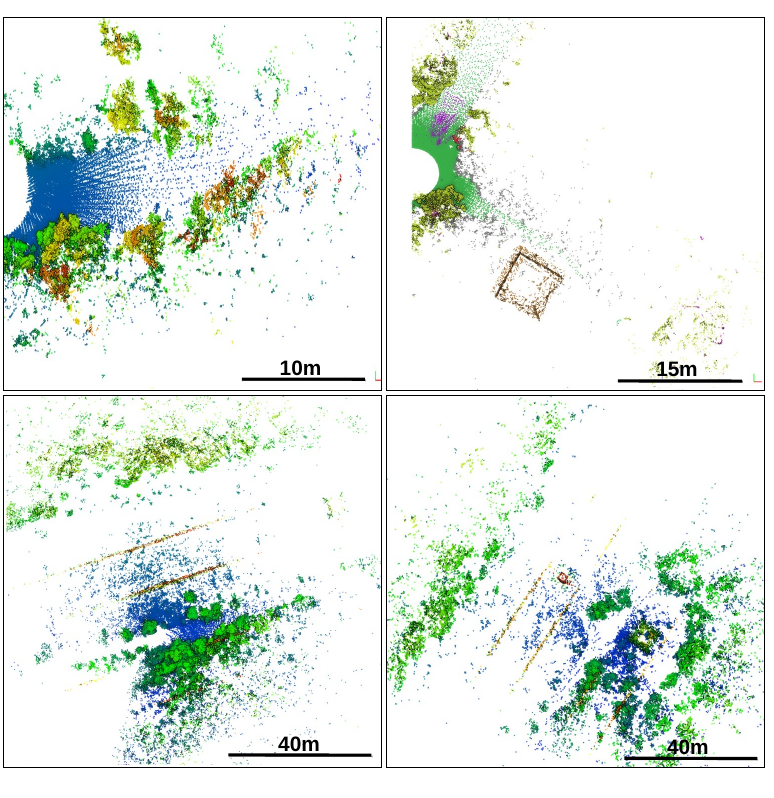}
    \caption{Examples of semantic submaps (top) compared to their corresponding submaps used for lidar PR (bottom). Semantic point clouds include only points falling in the frustum of the front camera. 
    }
    \label{fig:pcl_compare}
\end{figure}

\textbf{Multi-modality Place Recognition:}
We leverage the image-lidar pairs of WildScenes \cite{vidanapathirana2023wildscenes} to evaluate multi-modality place recognition in natural environments. Using the split generation described in WildScenes for semantic tasks, we split our data into train/val/test splits with $5990/283/2109$ matched pairs of images and point clouds, for a split ratio of roughly $70\%,5\%,25\%$. 3D submaps are voxel downsampled to 20K.  %
2D images are downsampled to 320x200 resolution.

We use MinkLoc++ \cite{komorowski2021minklocplus} for multi-modality place recognition as a baseline task. 
We only consider the inter-sequence place recognition problem in this experiment. A submap is considered successfully localised if the retrieved submap is within $5m$ of the query. During training, we construct triplets by selecting positive pairs within a $5m$ threshold of the query and negative pairs further than $50m$ from the query.

Evaluation results of MinkLoc++ on WildScenes are shown in Table \ref{tab:multimodal}. Interestingly, MinkLoc++ performs better on average when trained using only 3D modality, compared with training on multi-modal data. This could potentially be attributed to aliasing caused by repetitive patterns observed in forest environment images. However, this warrants further investigation, which we identify as a direction for future research.

\begin{table}[t]
    \centering
    \huge
    \caption{Multi-modality Place Recognition on WildScenes}
    \resizebox{1.0\linewidth}{!}{
    \begin{tabular}{cccccccc}
        \hline 
        \textbf{Method} & \textbf{Modality} &\multicolumn{2}{c}{\textbf{Venman}} & \multicolumn{2}{c}{\textbf{Karawatha}} & \multicolumn{2}{c}{\textbf{Average}}\\
        \cmidrule(lr){1-2}\cmidrule(lr){3-4} \cmidrule(lr){5-6} \cmidrule(lr){7-8}
        &&\textbf{R@1} & \textbf{R@1\%} & \textbf{R@1} & \textbf{R@1\%} & \textbf{R@1} & \textbf{R@1\%} \\
        MinkLoc++ & 3D + RGB & 32.9 & 75.2 & 45.4 & 89.8 & 39.2 & 82.5\\
        MinkLoc++ & 3D & 41.9 & 83.2 & 44.0 & 86.7 & 43.0 & 85.0\\
        
        \hline
    \end{tabular}
    }
    \label{tab:multimodal}
\end{table}

\section{Discussion}

In this section, we briefly discuss how field robotics can be improved with the use of large multi-modal benchmarks such as WildScenes and WildPlaces. A key strength in these benchmarks is the unstructured nature of natural environments, which is especially pronounced because the walking route used included off-track sections such as walking through long grass, over hanging low branches \--- capabilities that are very difficult for existing robotic platforms. Therefore, by collecting data in areas that are too difficult to collect with a robot, we are able to train robotic systems on data that would otherwise be difficult to obtain in deployment.

To date, we have demonstrated the tasks of lidar place recognition, semantic segmentation and domain adaptation. In this paper, we demonstrate the task of multi-modality place recognition using a combination of RGB images and lidar scans. We observe that these tasks experience reduced performance during temporal changes such as those experienced in long-term robotics situations.

One interesting observation is that the place recognition performance of MinkLoc++ is lower than that of similar lidar PR methods evaluated in WildPlaces. Specifically, considering single modality (3D only), MinkLoc++ achieves a recall@1 of $41.9\%$ on the Venman sequence. This is in comparison to a recall@1 of $75.77\%$ in WildPlaces on the same environment. While we note that the submaps have different sampling locations between WildPlaces and WildScenes, a key difference is in the size of these submaps \--- submaps in WildPlaces have 300K points versus semantic submaps in WildScenes with 70K points. This is because submaps in WildScenes only contain points that are visible within any 2D image from the dataset and have an associated semantic label. Figure \ref{fig:pcl_compare} shows examples of semantic point clouds compared to their corresponding point clouds generated from Wildcat.

\begin{figure}[t]
    \centering
    \includegraphics[width=0.99\linewidth, trim=0cm 0cm 0cm 1.4cm,clip]{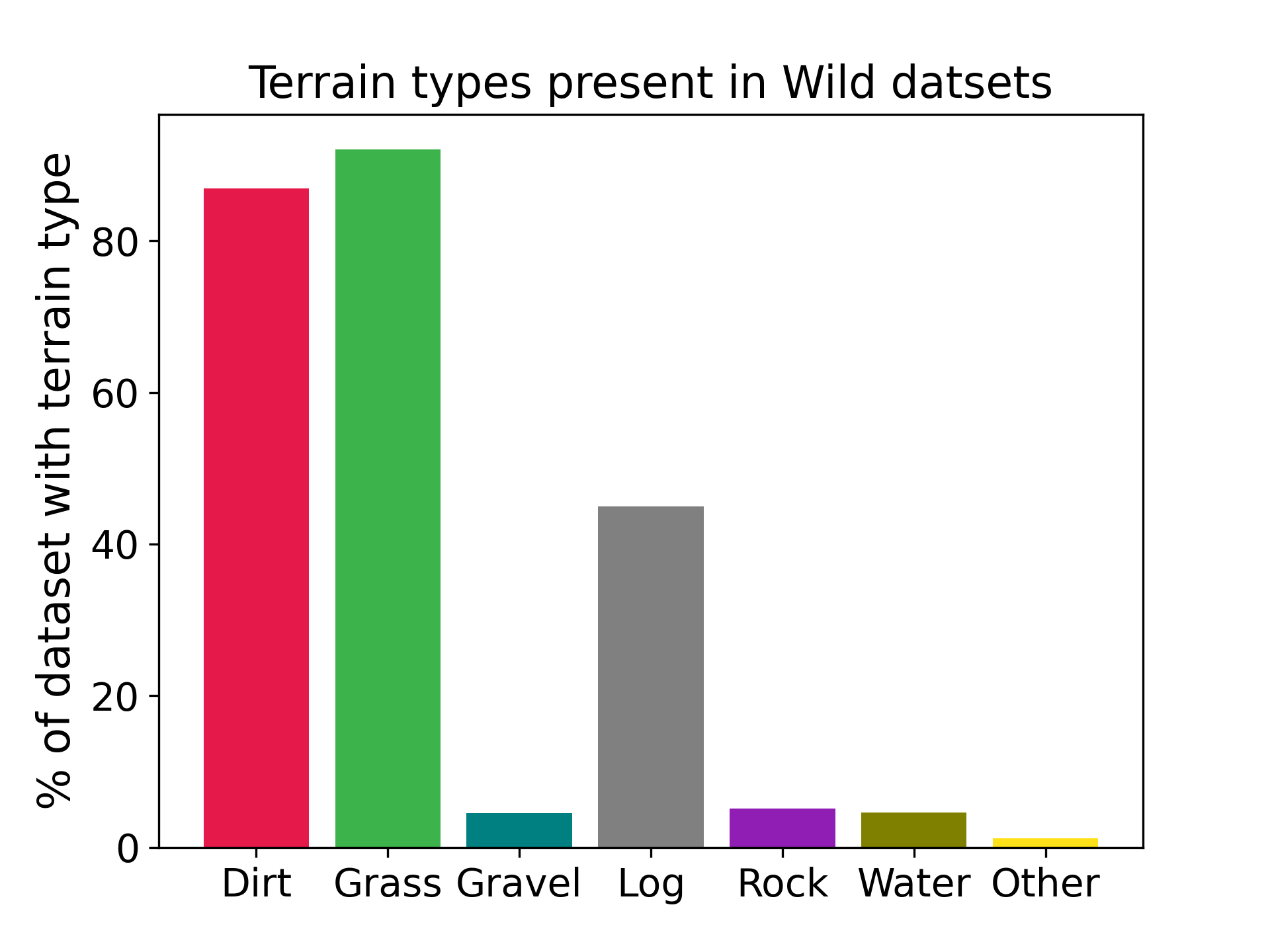}
    \caption{Terrain types present in our dataset.
    }
    \label{fig:trav}
\end{figure}

With the existing multi-modal dataset used in WildScenes, the data contains a lot of examples of difficult terrain which could be used for training neural networks for the task of traversability estimation. In Figure~\ref{fig:trav}, we provide a histogram of the percentage of different terrain types present in our data. By providing a diverse list of different terrain types, we believe our data would be suitable for training neural networks for traversability.

We demonstrated baselines for several tasks relevant to mobile robots including lidar place recognition, 2D semantic segmentation and 3D semantic segmentation. Future unexplored tasks include depth completion, traversability estimation, and optical flow.

\section{Conclusion and Future Work}

In this paper, we discussed potential use cases for the WildPlaces and WildScenes benchmarks in field robotics applications. We highlight that these benchmarks are useful in long-term robotic tasks including place recognition, 2D and 3D semantic segmentation, and domain adaptation tasks. 

For future work, it would be valuable to extend baselines to the traversability estimation, depth completion and optical flow tasks. Finally, we want to further investigate the development of multi-modality algorithms on our data by developing novel learning methods that can take in the full set of modalities available, that is, RGB images, lidar point clouds, and semantic labels.

\balance{}

\bibliographystyle{IEEEtran}
\bibliography{main}

\end{document}